\documentclass[letterpaper, 10 pt, conference]{ieeeconf}  

\IEEEoverridecommandlockouts                              

\usepackage{graphics} 
\usepackage{epsfig} 
\usepackage{mathptmx} 
\usepackage{times} 
\usepackage{amsmath} 
\usepackage{amssymb}  

\usepackage[utf8]{inputenc} 
\usepackage[T1]{fontenc}    
\usepackage{hyperref}       
\usepackage{url}            
\usepackage{booktabs}       
\usepackage{amsfonts}       
\usepackage{nicefrac}       
\usepackage{microtype}      

\usepackage{wrapfig}
\usepackage{cancel}
\usepackage{multirow}
\usepackage[font={footnotesize,it}]{caption}
\usepackage{graphicx}
\usepackage{pbox}
\usepackage{subcaption}
\usepackage[british]{babel}

\usepackage[normalem]{ulem}
\usepackage[dvipsnames]{xcolor}
\newcommand{\blazej}[1]{}
\newcommand{\todo}[1]{}
\newcommand{\pmilos}[1]{}
\newcommand{\negativeskip}{\vspace{-15pt}}
\newcommand{\tablenegativeskip}{\vspace{-8pt}}

\title{\LARGE \bf
CARLA Real Traffic Scenarios -- novel training ground and benchmark for autonomous driving
}

\author{Błażej Osiński$^{2,*,\$}$, Piotr Miłoś$^{1,3}$, Adam Jakubowski$^{\$}$, Paweł Zięcina$^{\$}$, Michał Martyniak$^{1}$, \\ Christopher Galias$^{4}$, Antonia Breuer$^{5}$, Silviu Homoceanu$^{5}$ and Henryk Michalewski$^{2,6}$%
  \thanks{*Correspondence to \texttt{b.osinski@mimuw.edu.pl}.}
\thanks{$^{1}$deepsense.ai, Warsaw, Poland, $^2$University of Warsaw, Warsaw, Poland}%
\thanks{$^3$Institute of Mathematics, Polish Academy of Sciences, Warsaw, Poland}
\thanks{$^4$Jagiellonian University, Kraków, Poland,$^5$Volkswagen AG, Wolfsburg, Germany, $^6$Google, London, UK, \$ - work done while at deepsense.ai}
}

\begin{document}

\maketitle
\thispagestyle{empty}
\pagestyle{empty}

\begin{abstract}
This work introduces interactive traffic scenarios in the CARLA simulator, which are based on real-world traffic. We concentrate on tactical tasks lasting several seconds, which are especially challenging for current control methods. The CARLA Real Traffic Scenarios (CRTS) is intended to be a training and testing ground for autonomous driving systems. To this end, we open-source the code under a permissive license and present a set of baseline policies. CRTS combines the realism of traffic scenarios and the flexibility of simulation. We use it to train agents using a reinforcement learning algorithm. We show how to obtain competitive policies and evaluate experimentally how observation types and reward schemes affect the training process and the resulting agent's behavior.
\end{abstract}


\section{Introduction}
\setcounter{footnote}{0}

The field of autonomous driving is a flourishing research area stimulated by the prospect of increasing safety and reducing demand for manual work. The ability to drive a car safely requires a diverse set of skills. First, it requires low-level control of a car -- being able to keep the lane, make turns, and perform emergency stops. It also requires an understanding of the traffic rules (such as speed limits, right of way, traffic lights) and adhering to them. Finally, it requires high-level navigation -- deciding which route to the destination is optimal. 

Computer-based systems can tackle these with varying levels of success.
High-level navigation is effectively solved by widely available GPS-based systems. There is also a growing body of work showing that the car control level can be efficiently solved using machine learning \cite{survey_tampuu, survey_kiran}.
Perhaps the hardest to automate is the middle layer of tactical planning in a time horizon of several seconds, which require adhering to traffic rules and reacting to other traffic participants.
In this work, we focus on \textit{changing lanes on highways} and \textit{driving through roundabouts}, facing interaction-intensive traffic not regulated by traffic lights. These maneuvers require the aforementioned tactical planning as well as following basic traffic rules, such as the right of way.


\begin{figure}[!t]
\centering
\begin{subfigure}{.45\textwidth}
  \centering
  \includegraphics[width=.80\linewidth]{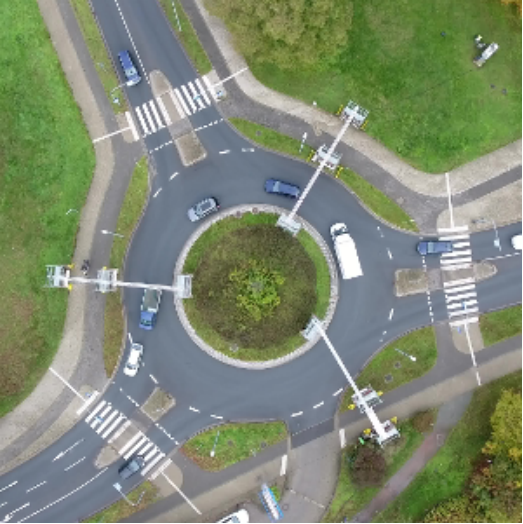}
\end{subfigure}
\begin{subfigure}{.45\textwidth}
  \centering
  \includegraphics[width=.80\linewidth]{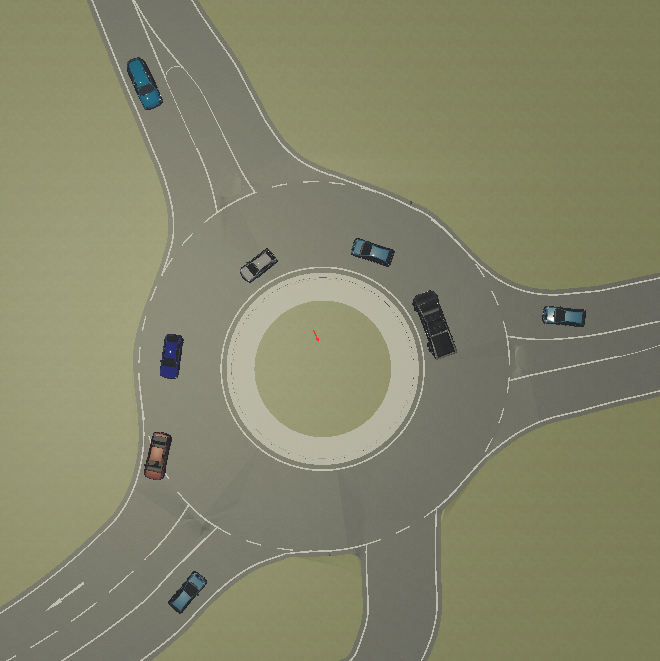}
\end{subfigure}
\caption{Illustration of real-to-sim approach applied in CRTS. A scene from openDD dataset together with its in-simulation counterpart.}
\label{fig:realtosim}
\negativeskip
\end{figure}

Simulators like \cite{carla, airsim, torcs} are popular within the autonomous research community. They greatly reduce the cost of data collection; however they introduce inherent imperfections known as the \emph{reality gap} \cite{jakobi1995realitygap}. We advocate that the gap can be partially mitigated by providing scenarios based on relatively cheap bird's-eye view data. We present CARLA Real Traffic Scenarios (CRTS), a suite of more than 60k simulated scenes of short-term tactical maneuvers. The scenarios are extracted from real-world datasets \cite{ngsim,openDD}, which offers an important benefit of tackling realistic situations. CRTS is intended to be used for training and validation of autonomous driving systems. We release the source code and custom made maps under a permissive open-source license\footnote{\url{https://github.com/deepsense-ai/carla-real-traffic-scenarios}. We provide also a Python package producing bird-eye view observation mode, which could be useful in unrelated projects:  \url{https://github.com/deepsense-ai/carla-birdeye-view}.}. Please refer to the project website for videos: \url{https://sites.google.com/view/carla-real-traffic-scenarios/}.

To the best of our knowledge, this is the first publicly available work that adapts tactical-level real-world traffic data to a simulator with rich plausible physics at a large scale. This unique setup has a number of benefits. It allows inexpensive and flexible evaluation and comparison of different class of approaches and algorithms in an environment close the real-world: the setup can be used for a rule-based system, classical planning, machine learning planers and more. Moreover, thanks to leveraging CARLA simulation, one can experiment with different input modalities being able to answer questions like: is Lidar useful to end-to-end planner when making a lane change? Finally, due to the standardized split into training and test scenarios, CRTS can be used to study generalization of control methods,
which has recently become a popular research direction; see e.g.~\cite{generalization, rl_overfitting}. We argue that CRTS decreases the reality gap enough, so that it can be used for reliable algorithmic testing and proof of concept experiments before real-world deployments.

Our second contribution is a comprehensive set of reinforcement learning experiments using CRTS, which illustrates the framework's versatility and serves as a reference point for the community.
In particular, we compare how various observation modes (bird's-eye view, visual input, and Lidar and camera) or reward structure affect performance of resulting policies. The results indicate very different capabilities of the policies to generalize, suggesting an interesting area for further study.


\section{Related work}
\label{sec:related_work}


\subsection{Methods for autonomous driving}

Autonomous driving is one of most important topics in contemporary AI research, with potential for a major societal impact. We refer to \cite{survey_kiran, Sadigh-RSS-16, Yurtsever2020} for general surveys.

Even for a seemingly narrow use-case as the lane change plenty of methods have been proposed. The classical planning methods are discussed in the survey \cite{ieee_survey}. From the more modern approaches,
\cite{ieee_rl_lane_change} uses a deep Q-network with continuous action and state spaces for automated lane change maneuvers, while \cite{nips_rl_lane_change} proposes a new state representation and utilizes an asynchronous DQN architecture in a multi-agent setting. 

\subsubsection{Deep learning}
Methods using deep neural networks (DNN) and learning have become quite fruitful for autonomous driving. There are three major paradigms of building DNN-based solutions (which are partially analogous to the discussion in cognitive sciences). The first one, \textit{mediated perception}, is perhaps the most prominent today; see e.g. \cite{ullman_1980,kitti,chauffeurnet}. It advocates using a vast number of sub-components recognizing all relevant objects (other cars, traffic lights, etc.) to form a comprehensive representation of the scene, which is then fed into a deep neural network. On the other side of the spectrum is the \textit{behavior reflex} approach, which suggests learning a direct mapping from raw observations to control actions. This approach is sometimes referred to as \emph{end-to-end}. It can be traced back to the 1980s \cite{Pomerleau88}, but has been growing in popularity lately as well \cite{BojarskiTDFFGJM16,wayverl,codevilla2018end,sim2real_icra}. The third paradigm, \textit{direct perception}, lies in the middle. It recommends learning affordances (e.g. the angle of the car relative to the road), which form a compact representation, which can then be used by another DNN to control the car; see e.g. \cite{DeepDriving2015,sauer2018conditional}.

\subsection{AV datasets}
On the other hand,
autonomous driving companies amass ever-increasing datasets of driving data collected in the wild. There is also an increasing number of publicly available datasets. A non-exhaustive list includes NGSIM \cite{ngsim} and highD \cite{highDdataset} datasets recorded on highways. inD \cite{inDDataset} and openDD \cite{openDD} concentrate on maneuvers on intersections and roundabouts. KITTI \cite{kitti}, A2D2 \cite{audi2020}, and \cite{lyft2020} datasets consist of footage recorded from cars using various sensors (RGB cameras, lidar) along with various preprocessing. SDD \cite{RobicquetSAS16} and CITR \cite{YangLRO19} put emphasis on humans and human-car interactions.

\subsection{Simulators}

Over the years, simulations have become a routinely-used tool for studying real-world control problems, including autonomous driving. Their clear advantages include making data collection cheaper and alleviating safety issues. The two popular packages CARLA \cite{carla} and AIRSIM \cite{airsim} use Unreal engine to build efficient photo-realistic simulations with plausible physics. SUMO \cite{lopez2018sumo} can simulate large traffic networks, but the visuals are not as realistic. Yet another example is TORCS \cite{torcs}, a widely-used racing car simulator. There are also other simulators which are attractive for autonomous driving research; we refer to \cite{simulators_review, chao2020survey} for comprehensive surveys.

\subsection{Simulated scenarios}
Apart from simulators, simulated scenarios have been proposed. For example in \cite{codevilla2019exploring} a new, more challenging benchmark \emph{NoCrash} was added to CARLA. Our work sits in the same context.

CARLA autonomous driving leaderboard at \url{https://leaderboard.carla.org/} presents a similar idea. There are at least two notable differences compared to our work: the leaderboard is intended only as a benchmark, without the possibility to be used for training, and it is also not based on real-life data.

Another interesting example is SMARTS \cite{zhou2020smarts}, which focuses on introducing highly interactive, artificial scenarios to SUMO. An notable case which did use real-world data is a work from Waymo \cite{scanlon2021waymo}, where they replay accidents in the simulator and show that their AV system would react better than human drivers. The scale of the experiment is however very small - it only contains 72 scenarios. 



\section{CARLA Real Traffic Scenarios (CRTS)} \label{sec:crts}
In this work we build interactive, realistic simulation-based scenarios of tactic maneuvers. Importantly, they are based on real-world data.
This is achieved by importing two datasets constructed from bird's-eye view video footage. NGSIM \cite{ngsimofficial} with highway traffic and openDD \cite{openDD} with roundabout data. From these datasets we extracted maneuvers of lane change and driving through a roundabout respectively; please refer to Table~\ref{table:CRTS-stats} for the number of extracted scenarios. To revive these situations in the CARLA simulator we have manually
built $9$ custom maps reflecting the areas from the datasets.
Following is the outline of creating an interactive scenario, which is further detailed in the subsequent sections.
We replace a car performing a maneuver with the ego vehicle simulated by CARLA and externally controlled (e.g. by an ML module) in a closed-loop manner.
The other vehicles follow the trajectory recorded in the dataset, see Section~\ref{sec:interactivity} for discussion on this design decision. The ego vehicle is tasked to execute the same maneuver. Performance is measured based on the final output (whether the ego-vehicle reached its target location) and negative events during driving (e.g. crashes, rapid turns, changes in speed). In particular, as we are not using an imitation reward, the policy steering the ego vehicle can use strategies different from the recorded to perform maneuvers.
The scenarios can be accessed using the standard OpenAI Gym interface \cite{openAiGym}.

Drawing inspiration from \cite{codevilla2018end}, in order to indicate the lane change direction or when to exit a roundabout the ego vehicle is provided a high-level navigational command (either \texttt{GO\_STRAIGHT}, \texttt{TURN\_RIGHT}, or \texttt{TURN\_LEFT}).

\begin{table}[h]
\resizebox{0.48\textwidth}{!}{%
\begin{tabular}{ c | c | c | c | c }
\toprule
\multirow{2}{*}{Source dataset} & \multirow{2}{*}{Maneuver type} & Our custom & \multicolumn{2}{c}{No. of scenarios}\\
 &  & maps & train & validation\\
\midrule
NGSIM  & highway lane change & 2 & 1750 & 467 \\
openDD  & drive through a roundabout & 7 & 51752 & 12927 \\
\bottomrule
\end{tabular}}
\caption{CARLA Real Traffic Scenarios (CRTS).}\label{table:CRTS-stats}
\tablenegativeskip
\end{table}

\subsection{Datasets and import procedure}

\subsubsection{Scenario extraction algorithm} \label{sec:scenario_extraction_algorithm}

In both cases of NGSIM and openDD, the datasets are scanned for maneuver events -- lane change and driving through roundabouts, respectively. The car performing the maneuver is declared to be the ego vehicle and a scenario is formed by mapping all traffic participants except for the ego vehicle to CARLA in the following way. A vehicle appearing in the field of view for the first time is spawned within the location and velocity matching the dataset. The model of vehicle is chosen from the CARLA library to match the dimension of the replayed car (measured using the Jaccard similarity). Later, consistency with the dataset is enforced every $100$ms (by setting again the locations and velocities). 

The initial speed and position of the ego vehicle is determined in the same way and in subsequent frames it is simulated by the CARLA physics engine according to the received actions. 

The set of such created scenarios is divided randomly into the train and test set in the $80$/$20$ ratio. 

\begin{figure*}[!t]
\begin{subfigure}{.19\textwidth}
  \centering
  \includegraphics[width=.85\linewidth]{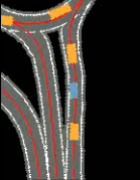}
  \caption{Bird's-eye view} \label{fig:obs_bev}
\end{subfigure}
\begin{subfigure}{.49\textwidth}
  \centering
  \includegraphics[width=.85\linewidth]{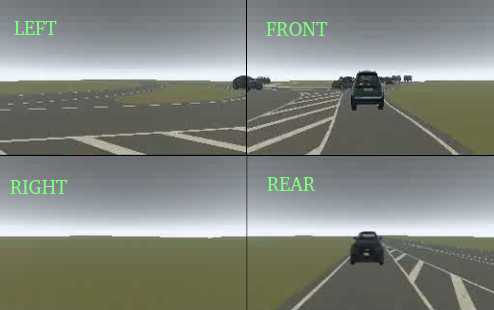}
  \caption{4 RGB cameras} \label{fig:obs_cam}
\end{subfigure}
\begin{subfigure}{.3\textwidth}
  \centering
  \includegraphics[width=.85\linewidth]{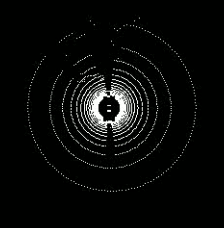}  
  \caption{Lidar's projection on 2D view} \label{fig:obs_lidar}
\end{subfigure}
\caption{Different observations types.}
\label{fig:possible_observations}
\negativeskip
\end{figure*}

\subsubsection{NGSIM}
In these scenarios, the ego vehicle is tasked with performing a lane change maneuver on a highway.
The source of the data is The Next Generation Simulation (NGSIM) dataset \cite{ngsim, ngsimofficial}, which contains vehicles trajectories collected on American highways. The dataset consists of 90 minutes of recordings.
For parsing the data, we used code from \cite{PPUU}.
We developed two custom CARLA maps of the locations covered by the dataset (southbound US 101 in Los Angeles, CA, and eastbound I-80 in Emeryville, CA).

A lane change event is declared as changing the lane id corresponding to a given vehicle. The scenarios starts $5$ sec before the lane change event. The scenario is considered successful if, at least for $1$ sec, the ego vehicle is located less than $30$cm of the target lane and its yaw is smaller than $10$ degrees. The scenario is unsuccessful if either a collision occurs, the car leaves both the starting and target lane or there is a timeout of $10$ sec.  

From NGSIM dataset, we extracted over $2$k scenarios of lane change maneuvers and assigned $1750$ for training and $467$ for validation.

During each scenario the following navigation commands are issued: \texttt{LANE\_CHANGE\_(}$dir$\texttt{)} if the ego vehicle is on the starting lane, $dir \in \{{left}, {right}\}$ denoting the direction to the target lane, otherwise \texttt{LANE\_FOLLOW}.

For the NGSIM we remap the position of the rear and front axle of the vehicle (to match the CARLA convention). We also smooth positions using a $1.5$ sec window. 

\subsubsection{openDD}
The openDD dataset \cite{openDD} is a large-scale roundabout trajectory dataset recorded from a bird’s eye view.
It is the largest publicly available trajectory dataset recorded from a drone containing over $84$k trajectories distilled from $62$ hours of video data. The data was collected on $7$ roundabouts -- five located in Wolfsburg and two in Ingolstadt (both in Germany). We developed $7$ custom CARLA maps corresponding to these roundabouts, see example in Figure~\ref{fig:realtosim}.

The scenarios begin when the ego vehicle is approx. $20$ meters before the roundabout entry. The scenario is considered successful if the ego vehicle exits the roundabout via the same exit as the reference driver. The scenario is unsuccessful if either a collision occurs, the car moves away more than $3$m from the original trajectory or there is a timeout of $1.5$ times of the time of the original drive.

From the openDD dataset, we obtained over $64$k scenarios of roundabout crossing and assigned $51752$ for training and $12927$ for validation.

During the whole scenario \texttt{LANE\_FOLLOW} command is issued until the car passes the last exit before the target one. Then the command changes to \texttt{TURN\_RIGHT}.

The openDD dataset is recorded at $30$ fps, which we downsample to match our $10$ fps.

\subsection{CARLA configuration}
We use the recent version 0.9.9.4 of CARLA \cite{carla}.
In this work we fix the low-quality CARLA settings, set the standard weather, and choose the ego vehicle to be Audi A2.
These configurations can however be easily changed.

\subsection{Metrics and reward functions}\label{sec:rewards}
The primary metrics used in CRTS is the average success rate of the performed maneuver on all test scenarios. As our code is open, other custom metrics can be defined. One could, for example, imagine measuring the comfort of driving.

For the purpose of training reinforcement learning agents we also define reward functions. At the end of a scenario a reward of $1$ (resp. $-1$) is issued if the scenario is successful (resp. unsuccessful).
Additionally we extend it with a dense component:

\begin{itemize}
    \item for NGSIM: the distance to the target lane is segmented into $10$ pieces. A reward of $0.1$ is issued each time the car moves a segment closer to the target lane and is penalized with $-0.1$ if it moves to a more distant segment.
    \item for openDD: The original trajectory is segmented into $10$ pieces. The agent is rewarded with $0.1$ for passing each segment.
\end{itemize}

\subsection{Observation space}
\label{sec:observation_space}

One of the crucial advantages of CRTS is its flexibility due to the use of simulation. In particular, one can easily test various observation settings. In our experiments we provide three major options: \emph{bird's-eye view}, \emph{visual input}, and \emph{lidar and camera}; see examples in Figure \ref{fig:possible_observations}.

The \emph{bird's-eye view} input covers approximately  $47 m \times 38 m$ meters of physical space encoded in $186 \times 150$ pixels. Semantic information is contained in $5$ channels corresponding respectively to road, lanes, centerlines, other vehicles, and the ego vehicle. The importance of these different features is measured in Section~\ref{sec:BEV-experiments}.

In \emph{visual} experiments we use 4 cameras (front, left, right, and rear), each having a field of vision of $90$ degrees and a resolution of $384 \times 160$. In \textit{front camera} experiments we use only the front-facing camera.

In \emph{lidar and camera} experiments, we use a single front camera and $360^{\circ}$ lidar inputs.
The simulated lidar has a range of $80$ meters, $32$ channels, horizontal field of view between $-15^{\circ}$, and $+10^{\circ}$; these parameters resemble lidar settings available in autonomous vehicles (see e.g. \cite{audi2020}).
In order to be able to use a convolutional neural network to process the lidar input, the raw data (pointcloud) is projected to a 2D view; this is a popular approach (see \cite{Yurtsever2020}).

\subsection{Action space}
The ego vehicle is controlled by steering and speed. The steering wheel's angle can be set continuously and is processed by the CARLA engine to steer the wheel (the angle of the wheels can vary from $-80^{\circ}$ to $80^{\circ}$, which is normalized to $[-1, 1]$ for the policy output). The speed is controlled indirectly by issuing commands to a PID controller, which controls the throttle and break.
For both steering and speed, the policy is modeled using the Gaussian distribution with the mean and the logarithm of the variance output by a neural network.

\begin{table*}[h]
\centering
\resizebox{\textwidth}{!}{
\begin{tabular}{l|ccc|ccc}
\toprule
{} & \multicolumn{3}{c|}{Base experiments} & \multicolumn{3}{c}{Bird's-eye ablations}  \\
{} & bird's-eye & visual & lidar \& camera & front only & no centerline & framestack   \\
\midrule
ngsim  &        $0.787 \pm 0.029$ &     $0.770 \pm 0.022$ &         $0.776 \pm 0.014$ &   $0.782 \pm 0.035$ &      $0.782 \pm 0.016$ &   $0.805 \pm 0.022$ \\
opendd &        $0.862 \pm 0.037$ &     $0.886 \pm 0.023$ &         $0.805 \pm 0.106$ &   $0.824 \pm 0.058$ &      $0.833 \pm 0.029$ &   $0.762 \pm 0.083$ \\
\bottomrule
\end{tabular}}
\caption{Means and standard deviations of success rates of various models, obtained over three experiments. For each experiment success rate is calculated on the same scenarios from the validation set.} \label{table:generalization}
\end{table*}

\subsection{Interactivity}
\label{sec:interactivity}
In the current version of CRTS non-ego agents are following trajectories recorded in the dataset. This design decision, sometimes referred to as data-replay, might seem controversial, but it has significant benefits over the alternatives, which include hand-crafted heuristics and a learned-based approach.\pmilos{Sentence is too long. Perhaps remove the last part and put (discussed below)}
Firstly, such a setup always allows for a successful completion of the maneuver -- it is enough for the ego vehicle to approximately follow the recorded trajectory. Moreover, the alternatives to replaying agents trajectories have serious drawbacks, as discussed in \cite{PPUU} (which also used data-replay for non-ego agents). The first alternative, using hand-crafted heuristics, is unlikely to cover the full diversity of drivers’ behaviour. The second alternative, the learned-based approach, faces the cold start problem, as we need realistic agents to train realistic agents. Even if we train the agents using imitation learning from recorded data (similarly to \cite{bergamini2021simnet}), it is not clear how they respond to novel 
behaviors of the ego vehicle, which would be out of their training distribution. Finally, having „over-reactive” agents, always maximizing safety even beyond the abilities of a human driver, might lead to the ego taking advantage of them - e.g. in the case of lane change, the best strategy might be changing it immediately, with disregard of others as they will always yield way. For these reasons, extending CRTS with realistic interactive agents seems like an interesting, but challenging future work. A potential idea is integration with SMARTS \cite{zhou2020smarts}. This project promises to collect and accumulate realistically behaving agents. At the moment however, even the definition of a metric for realism is not obvious.

\subsection{Generalization}
Recently it was identified that many of the common RL benchmarks are prone to overfitting, as they use the same version of environment during train and test time \cite{generalization, rl_overfitting}. CRTS was built to avoid this problem, as it contains separate suits of training and testing scenarios. This might make it an interesting environment for fundamental reinforcement learning research, not only in autonomous driving.

The experimental results presented in Section~\ref{sec:experiments} indicate that generalization is affected both by input modality, but also by reward shaping. This fact raises an interesting research question, how to design reinforcement learning algorithms for which generalization would not be impacted by these factors.


\section{Experiments}\label{sec:experiments}

Our experiments' main aim was to show how CRTS can be used in a versatile way to answer research questions concerning autonomous driving, sometimes arriving at counter-intuitive conclusions. The results are intended to be baselines for other researchers who would like to utilize CRTS.

The specific research questions are: a) can reinforcement learning be used to obtain maneuvering policies perform on test scenarios? b) how does the observation mode and reward structure affect the training and the quality of resulting polices? c) how do the trained maneuvering strategies compare to realistic (human) ones?

In our experiments we used a slightly modified PPO algorithm \cite{ppo}.
PPO is one of the most popular model-free RL algorithms known for its flexibility and stability. Being an on-policy algorithm, it usually requires a substantial number of training samples. Importantly, using a simulator mitigates this issue.

Table~\ref{table:generalization} presents the success rate on the test set. Our key conclusions are that the bird's-eye view generalizes almost perfectly and is considerably better than the other observation modalities and that dense rewards are important to obtain good generalization. Details are provided in the subsequent sections.

\subsection{Comparison of various observation modes}
\label{sec:analysis}
As mentioned before, using a simulator makes it easy to test the performance of algorithms with different possible inputs.
Here we describe an experiment with the three default modalities supported by CRTS and detailed in Section~\ref{sec:observation_space}: bird's-eye view, visual, and Lidar \& camera.

During training, all three modalities showed similar performance. Based on the Table~\ref{table:generalization} we can also notice that for the NGSIM test dataset, the difference between models seems to be relatively small.
However, there is a difference in performance on the test set of openDD, where Lidar \& camera is noticeably worse than the other two. This can be explained by the fact that Lidar and front camera only might provide too little data on the road curvature, which is crucial for effectively driving on roundabouts.




\subsection{Bird's-eye view experiments}
\label{sec:BEV-experiments}
Bird's-eye view, see Section~\ref{sec:observation_space} and Figure~\ref{fig:obs_bev}, is a top-down view of the road situation. It also resembles information that could be extracted from a perception system and HD-maps in a typical AV stack; see e.g. \cite{BreuerEJBHF19}.

\begin{figure}
\vspace{-20pt}
\includegraphics[width=0.5\textwidth]{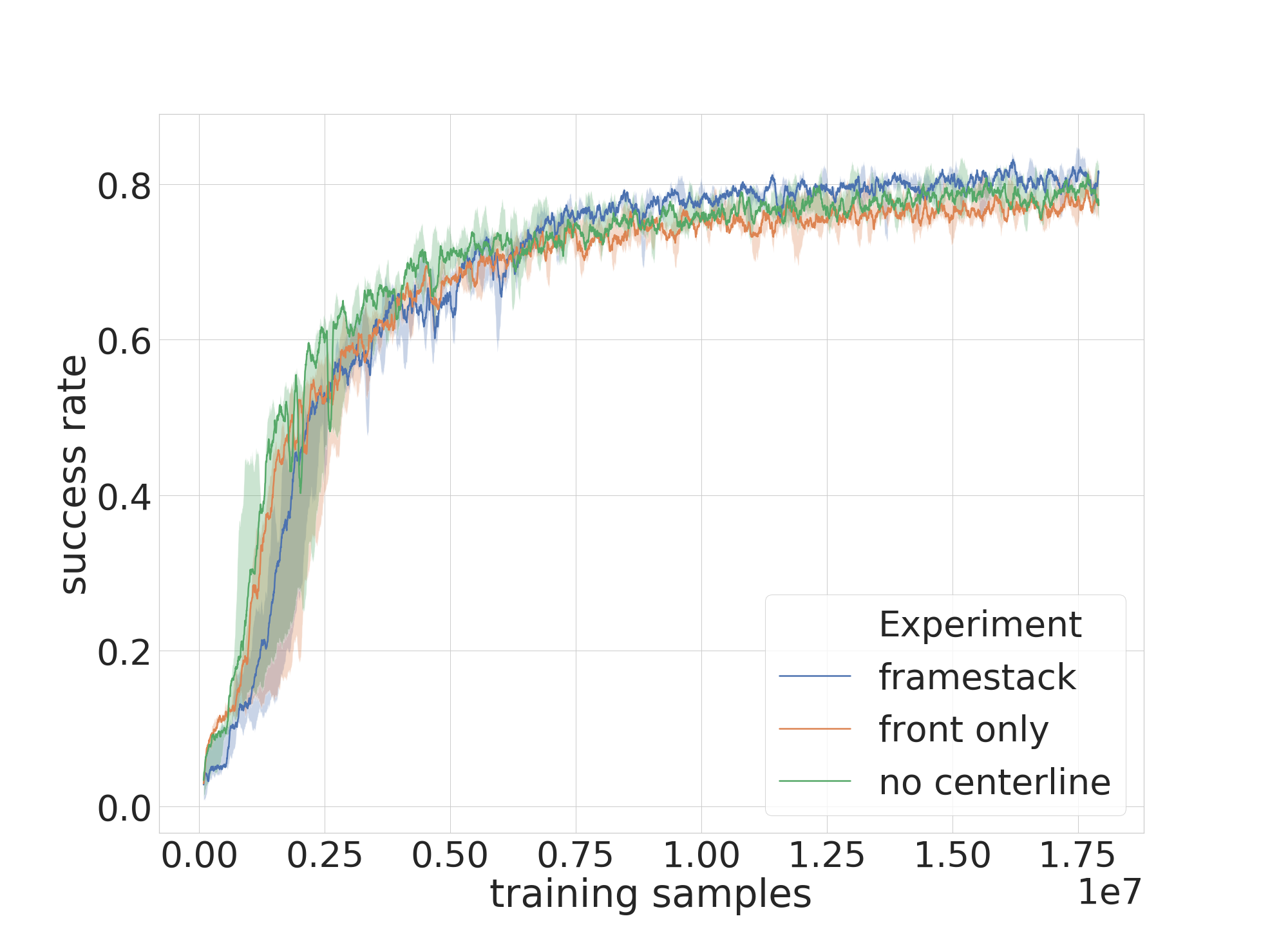}
\includegraphics[height=0.20\textwidth]{figures/abl_normal.png}
\includegraphics[height=0.20\textwidth]{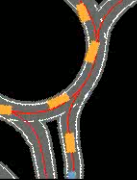}
\includegraphics[height=0.20\textwidth]{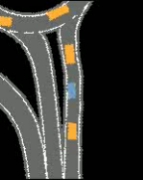}
\caption{Above: training curve for different variants of bird's-eye view comparisons on NGSIM. In each case three experiments were run. \\
Below: the same situation from openDD in three views. From left to right: base, front only, no centerline.}
\label{fig:birdeye-view-comparisons}
\vspace{-15pt}
\end{figure}

The observation in all \textit{bird's-eye view} baseline experiments is a tensor of shape $(186, 150, 5)$, which corresponds to a real-world area of approximately $47 m \times 38 m$. An interesting question is which components of the bird's-eye facilitate training. Along with the baseline experiment we ran the following modifications (visualized in Fig~\ref{fig:birdeye-view-comparisons}): \textit{front only} (covering only area in front of the ego), \textit{no centerline} (removing centerline channel), and \textit{frame stack} (stacking $4$ consecutive observations into $(186, 150, 20)$ tensor).

The performance of \emph{front only} and \emph{no centerline} are very close to each other on both datasets, during training and test. This is somehow surprising, because the \emph{front only} should struggle with doing a proper lane change maneuver without observing if an agent is approaching on the target lane. Evaluating the qualitative examples of the policy (which are also provided on the website), one can observe that the agent has found a strategy of immediately making a lane change and retreating if the lane is busy. This strategy does not always work, but is still surprisingly successful.

Comparing the results of base input with framestack gives interesting insight: framestack model is better on NGSIM, but noticeably worse on openDD. The first fact has a potential easy explanation - when the model has access to the last few frames, it can infer the speed of other agents, which is a useful feature for conducting lane change on a highway. The weaker performance on openDD might mean that estimating speed of other agents is not that useful on roundabouts. Another explanation, supported by qualitative examples in which the model drives too quickly and does not slow down, might be a case of causal confusion \cite{dehaan2019causal}, where the agent learned to maintain constant speed (which it can infer from the stacked frames).

\subsection{Reward experiments} \label{sec:reward_experiments}
In all the previous experiments, we used the \textit{dense reward} scheme described in Section~\ref{sec:rewards}. Here we show a comparison with two other schemes: \textit{sparse} and \textit{no failure penalty}. In the former, the agent gets a reward only at the end of the episode: $+1$ for the successful completion of a maneuver and $-1$ otherwise. The \textit{no failure} scheme is the same as dense except that the $-1$ failure penalty is not issued.

\begin{table}[h]
\begin{center}
    \begin{tabular}{l|ccc}
    \toprule
    {} & \multicolumn{3}{c}{Rewards experiments} \\
    {} & dense &  sparse & no failure penalty \\
    \midrule
    ngsim  &        0.828 &  0.741 &        0.707 \\
    opendd &        0.914 &  0.829 &        0.757 \\
    \bottomrule
    \end{tabular}
    \caption{Success rates on test scenarios for CRTS scenarios with bird's-eye view input.}\label{table:rewards}
    \tablenegativeskip
    \vspace{-4pt}
\end{center}
\end{table}

\begin{figure}
\includegraphics[width=0.50\textwidth]{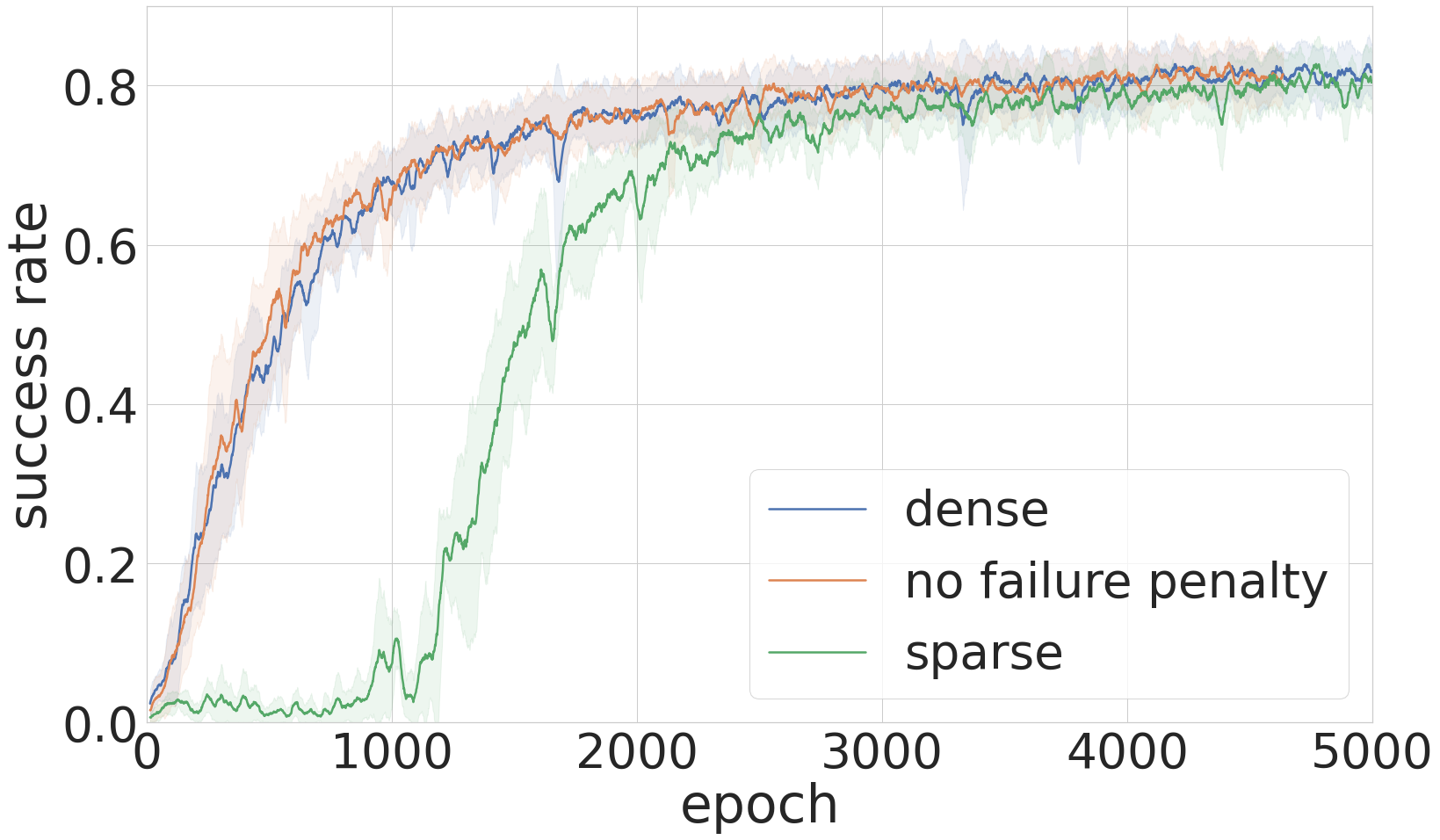}
\caption{Reward schemes comparison for NGSIM scenarios with bird's-eye view input.}\label{fig:reward_schemes_comparison}
\negativeskip
\end{figure}

Again we can compare the success rate from training (Figure~\ref{fig:reward_schemes_comparison}) and test (Table~\ref{table:rewards}).
The dense and no failure penalty look very similar during training, sparse reward takes more time to converge, but seems to reach the same performance.
In the evaluation however, dense reward is the clear winner. As the sparse reward is perhaps the most natural (the closest to a true metric of success rate), improving the RL algorithm to achieve the same test performance using this reward would be an interesting challenge.

\subsection{Qualitative analysis of driving styles}
There are many interesting questions concerning the obtained driving polices. Perhaps the most compelling one is asking if they are ``natural''. In an attempt to answer it we developed a tool for comparing the ego vehicle behavior with the original drive from the data-set; videos with qualitative examples of the policies are on the project website.
We observed that, in general, our policies drive much faster than the original agent. This aggressive driving style might be due to a lack of rewards for comfort and traffic rule following (we also observed a couple of situations of driving on the left lane). The quality of driving looks correct, although a few cases of avoidable crashes were observed. In the front-view experiments, there are rare instances of crashes with an unobserved rear vehicle.

\section{CONCLUSIONS AND FURTHER WORK} \label{sec:conclusions}
Our work introduces CRTS -- a new training and evaluation ground for autonomous driving based on real-world data. Our benchmark follows good practices of providing a train/test split. We hope that CRTS will serve as a platform for developing new methods and a sanity check before deployment in the real world. 

We provide benchmark evaluations using reinforcement learning, and observe that there is still much room for improvement. We provide a comparison between various observation settings and asses generalization and realism of obtained behaviors. 

There are many research directions to pursue further. Perhaps the most apparent is adding more scenarios and of different types (e.g. intersections from \cite{inDDataset}). Another direction is extending the metrics beyond success rate and introducing, for example, measures of comfort.
Finally, an interesting question stems from some of the reported experiments showing weaker generalization in other than baseline setups, like using sparse reward. What kind of RL algorithm could tackle these well?

Another research avenue is casting the problem into a multi-agent setup by making other driving actors controllable. An important new challenge, in this case, will be maintaining behaviors resembling real-world ones.


\section*{ACKNOWLEDGMENT}
We greatly acknowledge support of MathWorks which provided us with a free licence of RoadRunner. We used RoadRunner to prepare new CARLA maps.

The work of Piotr Miłoś was supported by the Polish National Science Center grants UMO-2017/26/E/ST6/00622.  This research was supported by the PL-Grid Infrastructure. We extensively used the Prometheus supercomputer, located in the Academic Computer Center Cyfronet in the AGH University of Science and Technology in Kraków, Poland. Our experiments were managed using https://neptune.ai. 
We thank the Neptune team for providing us access to the team version and technical support.

\addtolength{\textheight}{-0.7cm}   


\bibliographystyle{IEEEtran}
\bibliography{IEEEabrv,refs}

\end{document}